\documentclass[conference]{IEEEtran}
\IEEEoverridecommandlockouts
\usepackage{cite}
\usepackage{amsmath,amssymb,amsfonts}
\usepackage{algorithmic}
\usepackage{graphicx}
\usepackage{textcomp}
\usepackage{xcolor}
\usepackage{booktabs}
\usepackage{mytodonotes}
\usepackage{subcaption}
\usepackage[linesnumbered,ruled,vlined]{algorithm2e}
\usepackage[capitalise]{cleveref}
\usepackage[inline]{enumitem}

\newcommand{\approach}{SParSeFuL}

\def\BibTeX{{\rm B\kern-.05em{\sc i\kern-.025em b}\kern-.08em
    T\kern-.1667em\lower.7ex\hbox{E}\kern we demonstrated that \approach{} can reduce power consumption and (partially) communication overhead -.125emX}}
\begin{document}

% Some ideas for the title:
% Towards Sustainable Federated Learning: PSFL and Sparse Neural Networks for Large-Scale Systems
% Towards Sustainable Self-Federated Learning for Large-Scale Systems
% Towards Sustainable Self-Federated Learning
\newcommand{\meta}[1]{\textcolor{red}{#1}}
\title{Sparse Self-Federated Learning for Energy Efficient Cooperative Intelligence in Society 5.0}

\author{
\IEEEauthorblockN{Davide Domini}
\IEEEauthorblockA{%\textit{Department of Computer Science and Engineering} \\
\textit{University of Bologna}\\
Cesena, Italy \\
davide.domini@unibo.it}
\and
\IEEEauthorblockN{Laura Erhan}
\IEEEauthorblockA{%\textit{Faculty of Engineering} \\
\textit{Free University of Bozen}\\
Bolzano, Italy \\
Laura.Erhan@unibz.it}
\and
\IEEEauthorblockN{Gianluca Aguzzi}
\IEEEauthorblockA{%\textit{Department of Computer Science and Engineering} \\
\textit{University of Bologna}\\
Cesena, Italy \\
gianluca.aguzzi@unibo.it}
\and
\IEEEauthorblockN{Lucia Cavallaro}
\IEEEauthorblockA{%\textit{dept. name of organization (of Aff.)} \\
\textit{Radboud University}\\
Nijmegen, the Netherlands \\
lucia.cavallaro@ru.nl}
\and
\IEEEauthorblockN{Amirhossein Douzandeh Zenoozi}
\IEEEauthorblockA{%\textit{Faculty of Engineering} \\
\textit{Free University of Bozen}\\
Bolzano, Italy \\
adouzandehzenoozi@unibz.it}
\and
\IEEEauthorblockN{Antonio Liotta}
\IEEEauthorblockA{%\textit{Faculty of Engineering} \\
\textit{Free University of Bozen}\\
Bolzano, Italy \\
Antonio.Liotta@unibz.it}
\and
\IEEEauthorblockN{Mirko Viroli}
\IEEEauthorblockA{%\textit{Department of Computer Science and Engineering} \\
\textit{University of Bologna}\\
Cesena, Italy \\
mirko.viroli@unibo.it}
}

\maketitle

\begin{abstract}
Federated Learning offers privacy-preserving collaborative intelligence 
 but struggles to meet the sustainability demands of emerging IoT ecosystems 
 necessary for Society 5.0---a human-centered technological future balancing social advancement with environmental responsibility. 
The excessive communication bandwidth and computational resources required by traditional FL approaches
 make them environmentally unsustainable at scale, 
 creating a fundamental conflict 
 with green AI principles as billions of resource-constrained devices attempt to participate. To this end, 
 we introduce Sparse Proximity-based Self-Federated Learning (\approach{}), 
 a resource-aware approach that bridges this gap by 
 combining aggregate computing for self-organization with 
 neural network sparsification to %dramatically
 reduce energy and bandwidth consumption. 
% 
% Our experimental evaluation on the Extended MNIST dataset shows 
%  that \approach{} maintains competitive accuracy even with high sparsification ratios (up to 90\%).
%  % while reducing communication overhead by an equivalent percentage. 
% % 
% These results demonstrate that \approach{} enables environmentally sustainable federated learning for large-scale IoT ecosystems, 
%  advancing both Society 5.0's human-centered vision and green AI's imperative for resource efficiency in distributed machine learning.

\end{abstract}


\section{Introduction}\label{sec:intro}
\subsection{Context}
The Internet of Things (IoT) landscape is rapidly evolving, 
 with systems growing increasingly complex and ubiquitous. 
 This expansion necessitates novel paradigms capable of scaling to meet emerging societal challenges.
 Society 5.0~\cite{deguchi2020society}, 
 the vision of a human-centered society that balances economic advancement with social problem-solving,
 emphasizes a shift from centralized infrastructures toward decentralized, 
 self-organizing systems that can effectively manage the complexity of modern requirements.
 Central to this vision is environmental sustainability, 
 where technological innovation must operate within ecological constraints
 while still delivering societal benefits.

In this decentralized landscape, 
 machine learning plays a crucial role in enabling adaptive behavior across distributed agents. 
 However, traditional learning approaches often rely on centralized data collection, 
 raising significant privacy concerns and creating communication bottlenecks. 
 Furthermore, the escalating energy consumption of AI systems contradicts green computing principles,
 creating an unsustainable trajectory for large-scale IoT deployments.
 Cooperative learning frameworks~\cite{DBLP:conf/aaai/ConitzerO23} have emerged as a promising solution, 
 allowing systems to achieve high performance while maintaining decentralization.
Federated Learning (FL) represents one such approach, 
 enabling model training across distributed devices without requiring raw data sharing. 
 This paradigm preserves data privacy while facilitating collaborative intelligence across IoT ecosystems,
 though its resource-intensive nature presents energy efficiency challenges at scale.
 
\subsection{Research Gap}
Despite the advantages of Federated Learning, 
 applying it in the context of Society 5.0 and large-scale IoT systems presents significant challenges. 
 Traditional FL approaches rely on a centralized component for model aggregation, 
 which conflicts with the vision of truly decentralized, self-organizing systems. 
When deploying FL in large-scale IoT environments, 
 several concerns emerge simultaneously: 
\begin{enumerate*}[label=(\roman*)]
 \item heterogeneous data distributions across devices (non-IID dataset); 
 \item resource constraints of edge devices for green AI; and 
 \item the absence of reliable centralized infrastructure. 
\end{enumerate*}
While various solutions address these challenges individually--clustering methods for 
 non-IID data~\cite{DBLP:journals/tit/GhoshCYR22,DBLP:journals/tbd/LiLLZSLZWW24}, compression techniques for resource constraints~\cite{DBLP:journals/iotj/PrakashDCQSCGP22}, 
 and decentralized architectures peer to peer solutions~\cite{DBLP:conf/acsos/Domini24,DBLP:journals/jpdc/HegedusDJ21}--few 
 approaches tackle all these aspects holistically. 
Furthermore, these solutions introduce significant communication and computational overhead, 
 particularly when scaling to large numbers of devices with heterogeneous capabilities. 
The exchange of full model parameters during federation requires substantial bandwidth, 
 while model training and aggregation demand considerable computational resources, 
 limiting their practical applicability in resource-constrained IoT environments 
 where energy efficiency and reduced latency are critical requirements.

\subsection{Contribution}
In this paper, 
 we introduce Sparse Proximity-based Self-Federated Learning (\approach{}), 
 a novel approach that holistically addresses the aforementioned challenges by combining 
 aggregate computing for self-organizing system management 
 with neural network compression techniques to reduce resource consumption. 
Our research addresses the critical question: 
\emph{
 How can we design a self-federated learning process capable of scaling 
 to large-scale IoT systems while maintaining high performance 
 and minimizing resource requirements?
}
Our contribution is threefold:
\begin{enumerate*}[label=(\roman*)]
    \item we extend the Proximity-based Self-Federated Learning paradigm~\cite{DBLP:conf/acsos/DominiAFVE24}
     by integrating neural network compression techniques (\textit{i.e.}, sparsification and quantization) 
     that reduce communication overhead, energy consumption and inference time;
    \item we propose a possible research roadmap to implement and evaluate the proposed approach; and
    \item we discuss the possible impacts and challenges of \approach{}.
\end{enumerate*}

\subsection{Outline}
The rest of the paper is organized as follows:
\Cref{sec:related} provides background and related works in federated learning, compression techniques and aggregate computing;
\Cref{sec:method} introduces the Sparse Proximity-based Self-Federated Learning approach;
\Cref{sec:roadmap} presents a roadmap for future works to implement and evaluate the proposed approach;
\Cref{sec:impact} discusses possible impacts of \approach{};
and \Cref{sec:conclusion} concludes the paper and outlines future research directions.

\section{Background and Related Works}\label{sec:related}

\subsection{Federated Learning}
Federated Learning (FL)~\cite{DBLP:conf/aistats/McMahanMRHA17} has emerged as a powerful framework to train machine learning 
 models in scenarios with stringent privacy concerns.
One of the most widely adopted approaches in FL is Federated Averaging (FedAvg)~\cite{DBLP:journals/corr/McMahanMRA16}, which enables
 decentralized model training without requiring raw data to be shared among devices. 
In recent years, FL has been increasingly applied in IoT scenarios, such as urban traffic  prediction~\cite{DBLP:conf/ijcai/Zhang00XLC24,DBLP:journals/corr/abs-2403-07444}, 
 where a large number of spatially distributed devices collaborate to learn a shared model while 
 preserving local data privacy.
In such IoT environments, several challenges naturally arise due to the inherent characteristics of the 
 distributed devices~\cite{DBLP:journals/iotj/ImteajTWLA22}. 
First, data heterogeneity is a significant issue~\cite{DBLP:conf/icde/LiDCH22}, as devices perceive different subportions of 
 the environment, leading to non-independently and identically distributed (non-IID) data. 
Second, many IoT devices operate under resource constraints, including limited computational power, memory, and energy availability, 
 which can hinder efficient model training and communication~\cite{DBLP:journals/iotj/PrakashDCQSCGP22,Imteaj2023}.

Various solutions have been proposed to address data heterogeneity. 
Some approaches, such as FedProx~\cite{DBLP:conf/mlsys/LiSZSTS20} and Scaffold~\cite{DBLP:conf/icml/KarimireddyKMRS20}, 
 introduce regularization terms into the loss function to mitigate local model drift and maintain alignment with the global objective. 
Other methods assume that clients can be partitioned into clusters based on data similarity, with each cluster containing clients 
 that experience independent and identically distributed (IID) data.
Notable examples of this clustering-based strategy include IFCA~\cite{DBLP:journals/tit/GhoshCYR22}, 
 PANM~\cite{DBLP:journals/tbd/LiLLZSLZWW24}, and FedSKA~\cite{DBLP:conf/ecai/Li0TWXZ23}.
In this study, we focus on a subset of clustering-based approaches that leverage spatial information to improve model aggregation. 
This choice is motivated by the fact that our application domain is large-scale IoT ecosystems in Society 5.0, where the spatial distribution 
 of devices plays a crucial role in the federated learning process.
Specifically, we investigate algorithms that assume the clustering of devices is influenced by their spatial proximity, 
 under the hypothesis that devices in close proximity are more likely to experience similar environmental conditions 
 than those farther apart~\cite{esterle2022deep,DBLP:conf/sac/MalucelliDAV25}. 
Among the methods in this category are Field-based Federated Learning (FBFL)~\cite{DBLP:conf/coordination/DominiAEV24} and 
 Proximity-Based Self Federated Learning (PSFL)~\cite{DBLP:conf/acsos/DominiAFVE24}---the one we focus on in this work. 

\subsection{Compression Techniques}\label{sec:compression}
Classic deep learning methods are usually computationally heavy and involve a communication overhead to transfer collected 
 data to powerful computers or the cloud and then process them further.
With the development of the computational capabilities of IoT devices, we are witnessing a more capable Edge and therefore, 
 a push towards leveraging the Edge fuelled by a strong desire for sustainability, energy efficiency, 
 minimizing communication overhead, bandwidth usage, and increased scalability~\cite{Li2020-xw}. 
Making use of the edge does not resume to running the heavy deep learning algorithms on smaller devices but is also concerned with investigating 
 the optimization of network architectures, and model compression through quantization and pruning techniques, to name a few~\cite{Liang2021-ta}. 

Sparse Artificial Neural Networks, \textit{i.e.}, ANNs with fewer connections (usually achieved via pruning techniques), 
 gained popularity in recent years as they can achieve performance levels comparable to those of their fully connected counterparts 
 while boasting reduced model size, a more lightweight ANN structure (reduced number of parameters and connections), and lower 
 inference times~\cite{Erhan2025-gi, Jayasimhan2024-mq, Atashgahi2024-ki}. This holds even for simple static pruning approaches, 
 such as pruning an ANN after training.

While pruning and sparsification are concerned with reducing the number of connections in an ANN according to some conditions, 
 quantization implies reducing the precision of the model's weights, gradients or activations to achieve a reduced memory footprint 
 and inference time while minimizing accuracy loss ~\cite{krishnamoorthi2018quantizingdeepconvolutionalnetworks}.

\subsection{Aggregate Computing}
Aggregate Computing (AC)~\cite{DBLP:journals/computer/BealPV15} is a macro-programming paradigm~\cite{DBLP:journals/csur/Casadei23} 
 that abstracts from individual device behaviors to focus on collective system dynamics. 
 AC provides a functional approach to distributed computing by expressing collective behaviors through 
 computational fields~\cite{DBLP:journals/pervasive/MameiZL04,DBLP:journals/tocl/AudritoVDPB19}—distributed data structures 
 that map computational devices to values across space and time.
This paradigm is particularly suitable for the resource-constrained IoT environments described in Society 5.0, 
 as it addresses scalability challenges by design, managing system evolution through field manipulations rather than 
 device-by-device coordination. 
The key benefits of AC in scenarios of interest include its spatio-temporal universality~\cite{DBLP:conf/coordination/AudritoBDV18} 
 (ability to model diverse distributed behaviors) and formal resilience guarantees, 
 such as self-stabilization~\cite{DBLP:journals/tomacs/ViroliABDP18} (recovery from perturbations) and 
 eventual consistency~\cite{DBLP:journals/taas/BealVPD17} (convergence regardless of network density). 
 These properties align perfectly with Society 5.0's vision of robust, self-organizing systems that can operate efficiently 
 at scale while minimizing resource consumption.

A key pattern in AC is the Self-Organizing Coordination Region (SCR)~\cite{DBLP:conf/coordination/CasadeiPVN19}, 
 which provides a formal framework for dynamically partitioning a network into contiguous \emph{regions}. 
In this context, a region represents a set of devices that collaboratively operate under the coordination of a designated leader,
 selected through a system-wide multi-leader election process.
The SCR pattern exhibits strong self-stabilization properties; when a leader fails or leaves the network, 
 the system autonomously reconfigures to maintain operational continuity and structural 
 integrity---making it particularly effective in federated learning for maintaining an 
 uninterrupted learning process~\cite{DBLP:journals/corr/abs-2502-08577}.
The information flow within each region follows a well-defined protocol:
\begin{enumerate*}[label=(\roman*)]
 \item local node data is propagated to the leader through a gradient-based multi-hop diffusion mechanism---see the reference paper for 
  more details of how it works~\cite{DBLP:conf/coordination/CasadeiPVN19};
 \item the leader performs aggregation computations to synthesize a comprehensive regional perspective; and
 \item based on this global view, the leader disseminates control signals throughout the region.
\end{enumerate*}
AC provides three fundamental self-organizing computational blocks that enable the implementation of the SCR pattern:
\begin{enumerate*}[label=(\roman*)]
    \item S-block for sparse-choice, distributed multi-leader election;
    \item G-block for gradient-based information propagation; and
    \item C-block for data collection and aggregation directed toward a designated sink device. 
\end{enumerate*}
The integration of AC principles with SCR patterns has demonstrated significant efficacy in distributed machine 
 learning contexts~\cite{DBLP:journals/scp/DominiCAV24,DBLP:conf/acsos/AguzziVE23}.
Our current work builds upon this established foundation to develop a novel, resource-efficient approach to federated learning.

\section{Sparse Proximity-based Self-Federated Learning}\label{sec:method}

\subsection{Problem Formalization}
\begin{figure}
    \centering
    \includegraphics[width=0.7\columnwidth]{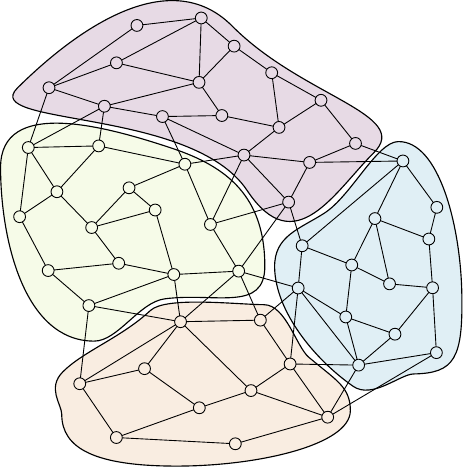}
    \caption{
    Visual representation of data skewness between different subregions. 
    In this case, an area $A$ is divided into four subregions.
    Each color represents a different data distribution. 
    Data within the same subregion are uniformly distributed, while data from different subregions exhibit non-IID properties.
    Each circle represents a different device, while the lines between devices represent communication links.
    }
    \label{fig:subregions}
\end{figure}

In this work we follow the formalization of Proximity-Based Federated Learning 
 given in~\cite{DBLP:conf/acsos/DominiAFVE24}.
In particular, we consider an area $A = \{a_1, \dots, a_k\}$ divided into $k$ distinct continuous 
 sub-regions~(\Cref{fig:subregions}).
Each sub-region $a_j$ has a unique local data distribution $\Theta_j$, thus resulting in a non-IID setting. A set of devices $S = \{s_1, \dots, s_n\}$ 
 (with $n \gg |A|$) is deployed in $A$, each capable of maintaining data and participating the federated learning process.
Each device is located in a specific area $a_j$, though it does not have access to this information.
Furthermore, each device can communicate only with a limited set of other devices (\textit{e.g.}, those within a specific 
 communication range $r_c$), thereby defining a neighborhood $N_i$ for each device $s_i$.

Locally, each device $s_i$ creates a dataset $D_i$ from samples drawn from the data distribution $\Theta_i$.
Herein, we consider a classification task where each sample $d_t$ in the data distribution $\Theta_i$
 consists of a feature vector $\mathbf{x}_t$ and a label $y_t$.
Therefore, the complete local dataset is represented as $D_i = \{(\mathbf{x}_1, y_1), \dots, (\mathbf{x}_m, y_m)\}$.

For each time step $t$, the nodes should devise a set of \emph{federations} $F^t = \{f_1^t, \dots, f_m^t\}$, 
 where each federation $f_j$ is represented as a tuple $(l_j, S_j) $, with:
\begin{enumerate*}[label=(\roman*)]
    \item $l_j$ the leader device responsible for the federation; and
    \item $S_j$ the set of nodes that are part of the federation.
\end{enumerate*}
Giving $T$ the number of time steps, the goal of our approach is to create a set of federations $F^T$ 
 which approximates the subregions $A$ and find the set of federation-wise models $\{\mathcal{M}_1^{T}, \dots, \mathcal{M}_m^{T}\}$
 (namely, one model for each federation).
Given a correct federations partitioning, we want to find a set of federation-wise models that minimize the 
 error across all areas, which can be formally described as:
\begin{equation}
    \min_{ \mathcal{M}^{T}} \sum_{\mathcal{M}_i^{T} \in \mathcal{M}^{T}} L(\Theta_i, \mathcal{M}_i^{T})
\end{equation}
Where $L$ represent the average error computed by a loss function from each sample $x_i$ and label $y_i$ in 
 the dataset $\Theta_i$ using the model $\mathcal{M}_i^{T}$.

\subsection{Proposed Approach}

\begin{figure*}
    \centering
    \includegraphics[width=1\textwidth]{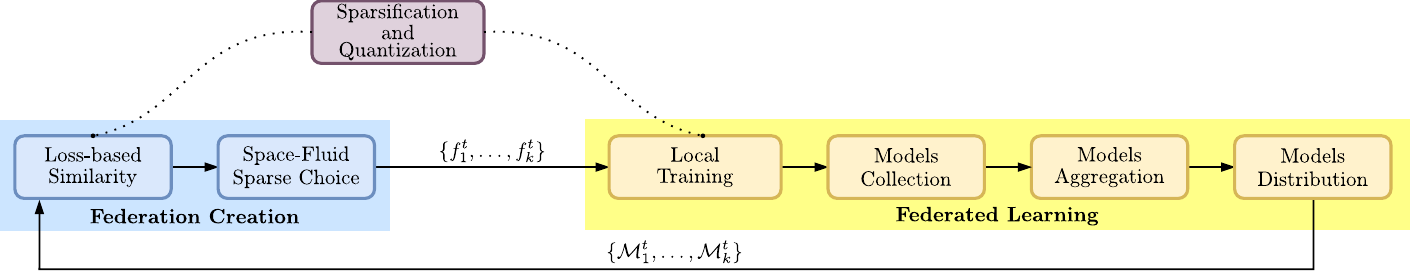}
    \caption{
      Overview of the proposed algorithm, with highlighted components where compression techniques 
      discussed in~\Cref{sec:compression} can be applied.
    }
    \label{fig:algorithm}
\end{figure*}

The approach introduced in this paper, \approach{} (see~\Cref{alg:self_fed_learning}), builds upon the Proximity-based Self-Federated 
 Learning (PSFL) framework.
Our primary contribution is the integration of model compression techniques into PSFL (\Cref{fig:algorithm}) 
 to enhance resource efficiency during the learning process.
The fundamental learning flow of PSFL consists of four key steps: 
\begin{enumerate}
    \item \emph{Federations Creation}: devices form a federation via a distributed multi-leader election process. 
     The elected leaders are responsible for the coordination and management of federation activities;
    \item \emph{Models Collection}: the leaders collect models from each device within the federation;
    \item \emph{Models Aggregation}: based on the aggregated models, the leaders synthesize a unified federation-wise model
     that reflects the collective learning outcomes of the federation; 
    \item \emph{Models Distribution}: the synthesized model is then disseminated by the leaders to all nodes in the federation,
     ensuring that each node operates with the most updated and accurate model reflective of their shared data environment.
\end{enumerate}

\begin{figure*}
    \centering
    \begin{subfigure}[b]{0.25\textwidth}
        \includegraphics[width=\linewidth]{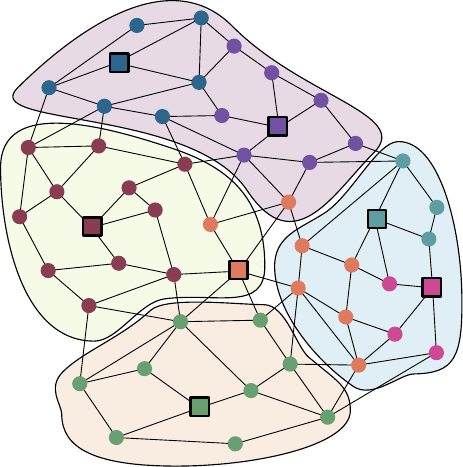}
    \end{subfigure}
    \begin{subfigure}[b]{0.25\textwidth}
        \includegraphics[width=\linewidth]{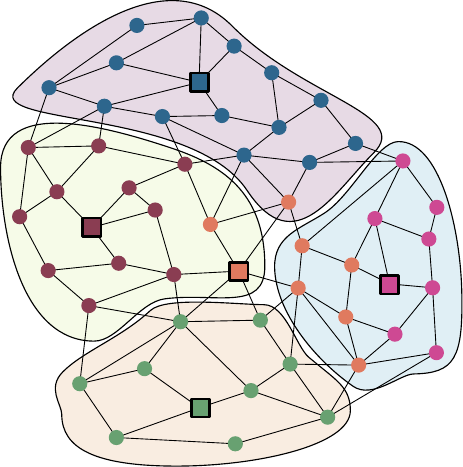}
    \end{subfigure}
    \begin{subfigure}[b]{0.25\textwidth}
        \includegraphics[width=\linewidth]{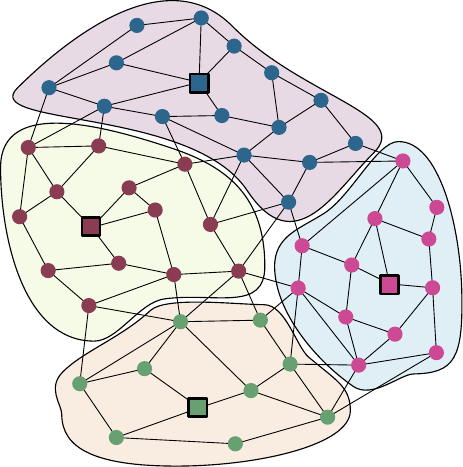}
    \end{subfigure}
    \caption{
       Graphical representation of federations evolution over time. 
       As in~\Cref{fig:subregions}, we have four different subregions (\textit{i.e.}, four heterogeneous data distributions).
       The color of each device represents the federation to which it belongs. 
       Squared devices are the leader of their federation.
       The time flows from left to right.
       It is possible to observe that, initially, the number of federations does not match the actual number of distinct data distributions.
       However, over time, it gradually converges to the correct value.
    }\label{fig:federations-evolution}
\end{figure*}

This learning process introduces a degree of dynamicity in the formation and evolution of federations over 
 time~(\Cref{fig:federations-evolution}).
This dynamic behavior arises from the fact that, at each global round, devices evaluate a similarity metric $ds$ relative 
 to their neighbors to determine whether they are experiencing comparable data distributions.
Due to this similarity-based check, the total number of federations may exhibit an oscillatory behavior throughout 
 training. 
However, under proper similarity approximation, federations eventually converge to the true number of distinct data distributions 
 present in the system. 
Notably, if the similarity check is based on a poor approximation, such as an excessively high sparsity level (\textit{e.g.}, $\psi \ge 0.8$), 
 the process may fail to properly differentiate distributions, leading to inaccurate federation formation.

A key challenge in performing similarity checks is that, due to privacy constraints, devices cannot directly compare 
 their local datasets. 
To address this, we employ a loss-based similarity metric as a proxy for data distribution similarity.
Specifically, given two devices $s_i$ and $s_j$ with their respective models $\mathcal{M}_i$ and $\mathcal{M}_j$,
 the similarity is computed as follow:
\begin{enumerate*}[label=(\roman*)]
    \item \emph{Models Exchange}: devices share their trained models;
    \item \emph{Cross Validation}: each device evaluates the received model on its own local dataset, 
     resulting in a loss value $L_{i,j}$ (\textit{i.e.}, the loss computed by $s_i$ using model $\mathcal{M}_j$) and 
     $L_{j,i}$ (similarly for device $s_j$ with model $\mathcal{M}_i$);
    \item \emph{Similarity Metric Computation}: the total similarity metric is obtained as the sum of both 
     losses $ds_{i,j} = L_{i,j} + L_{j,i}$;
    \item \emph{Federation Decision}: if the computed similarity metric is below a predefined threshold, the two devices 
     are considered to have sufficiently similar data distributions and can safely aggregate their models without 
     inducing model drift.
\end{enumerate*}

As illustrated in~\Cref{fig:algorithm}, \approach{} extends this pipeline by incorporating model compression techniques
 at critical stages to optimize efficiency.
Specifically, these techniques can be applied in two main components: 
\begin{enumerate*}[label=(\roman*)]
    \item during the \emph{similarity check} for federations creation, compressed models can be used to approximate similarity 
     instead of full models, reducing computational and communication costs; and
    \item during \emph{training} model compression can be leveraged to minimize storage and transmission overhead while maintaining 
     learning efficacy.
\end{enumerate*}

\begin{algorithm}
\caption{Sparse Proximity-based Self-Federated Learning (\approach{})}

\label{alg:self_fed_learning}

\DontPrintSemicolon
\SetKwInOut{Input}{Input}
\SetKwInOut{Output}{Output}

\Input{
 Set of devices $S = \{s_1, \ldots, s_n\}$, 
 each with initial model $\mathcal{M}_i^0$, 
 local dataset $\mathcal{D}_i$, 
 compression strategy $\mathcal{C}$,
 sparsification ratio $\psi \in [0,1]$,
 similarity threshold $\tau$}
\Output{Final trained models $\{\mathcal{M}_i^T\}_{i=1}^n$}

\For{each global round $t = 1,\ldots,T$}{
    \For{each device $s_i \in S$}{
    
        $\mathcal{M}_i^{t-1,\text{c}} \leftarrow \mathcal{C}(\mathcal{M}_i^{t-1}, \psi)$ 
    
        %\tcp{Train local model}
        $\mathcal{M}_i^{t,c} \leftarrow \text{LocalTraining}(\mathcal{M}_i^{t-1,c}, \mathcal{D}_i)$
        
        Broadcast $\mathcal{M}_i^{t,\text{c}}$ to neighborhood $\mathcal{N}_i$ \;
        
        Compute pairwise dissimilarity $d_{ij} = \mathcal{D}_s(\mathcal{M}_i^{t,\text{c}}, \mathcal{M}_j^{t,\text{c}})$ $\forall s_j \in \mathcal{N}_i$ \;
        
        \tcp{Apply SCR pattern for federation formation}
        
        Determine federation membership $\mathcal{F}_j$ using G-block based on $\{d_{ij}\}$ and threshold $\tau$ \;
        
        \If{$s_i$ is elected as leader via S-block}{
            $\{\mathcal{M}_k^{t,c}\}_{s_k \in \mathcal{F}_j} \leftarrow \text{CollectModels}(\mathcal{F}_j)$ via C-block \;
            
            $\mathcal{M}_{\mathcal{F}_j}^{t} \leftarrow \text{FedAvg}(\{\mathcal{M}_k^{t,c}\}_{s_k \in \mathcal{F}_j})$ 

            Disseminate $\mathcal{M}_{\mathcal{F}_j}^t$ to all devices in $\mathcal{F}_j$ via G-block \;
        }
        Update local model: $\mathcal{M}_i^{t+1} \leftarrow \mathcal{M}_{\mathcal{F}_j}^t$ 
    }
}
\end{algorithm}

\section{Research Roadmap}\label{sec:roadmap}
Establishing a coherent research roadmap necessitates a clear definition of our objectives. 
Broadly, these can be classified into two categories: \emph{functional} and \emph{non-functional} goals.
The functional objective primarily involves the development and implementation of cooperative learning strategies 
 to effectively solve a given task. 
This encompasses the design of mechanisms that enable agents to learn and adapt collaboratively.
Conversely, the non-functional objectives relate to performance and resource optimization. 
These include several key sub-goals, such as time, execution, memory, energy, and 
 communication efficiencies.
Addressing these aspects is crucial for ensuring the scalability and practical feasibility of our approach.

\subsection{Integrating Compression Techniques}

The first objective is the integration of the compression techniques outlined in~\Cref{sec:compression}. 
This process unfolds through several subgoals that define the degree of integration within our framework.
As illustrated in~\Cref{fig:algorithm}, sparsification and quantization can be incorporated into two key components of our approach: 
 the similarity check step and the learning process itself. 
A crucial first step involves integrating model compression within the similarity check step to approximate the similarity metric 
 used for federation formation.
Once it is verified that this approximation enables effective federation formation without degrading accuracy--or by identifying a suitable tradeoff--it 
 will be necessary to extend compression to the learning phase. 
This step must ensure that accuracy remains stable, avoiding significant performance drops.
Additionally, a comparative analysis of different sparsification strategies is required, distinguishing between static approaches 
 (\textit{e.g.}, pre- and post-pruning) and dynamic techniques (\textit{e.g.}, sparse evolutionary training~\cite{mocanu2018scalable}), 
 which adapt the network structure during learning.

\subsection{Real World Deployment}
After successfully integrating and validating the proposed compression techniques, 
 a crucial next step is the identification and implementation of real-world use cases. 
While the current evaluation primarily relies on well-known datasets--such as MNIST~\cite{lecun2010mnist}
 and CIFAR-10~\cite{cifar10}--split synthetically 
 to simulate federated learning scenarios, transitioning towards realistic applications is essential for demonstrating 
 the practical viability of our approach.

In this context, two particularly relevant application domains emerge: 
 traffic flow control and building heating optimization in smart cities. 
In the former, decentralized learning can optimize traffic management by enabling edge devices--such as cameras, sensors, 
 and connected vehicles--to collaboratively learn congestion patterns, predict bottlenecks, and adjust signal timing efficiently, 
 all while reducing communication overhead through model compression. 
Similarly, in the latter, intelligent heating control relies on distributed learning across multiple sensors to dynamically 
 adjust temperature settings and minimize energy consumption. 
A key characteristic shared by both scenarios is the spatial structure of data distributions: 
  devices in close proximity tend to perceive similar environmental conditions and thus generate correlated data. 
This phenomenon arises naturally in urban settings, where different neighborhoods exhibit distinct characteristics. 
For instance, in a city, the heating requirements of an industrial district will differ significantly from those of 
 a residential or university area.

By transitioning from synthetic benchmarks to such real-world applications, we can assess not only the accuracy and efficiency 
 of our methods but also their scalability and adaptability in complex, dynamic environments.

\subsection{Self-Adaptive Compression}
Another promising research direction is the development of self-adaptive compression parameters for neural networks, tailored to the conditions 
 perceived by the devices. 
This approach enables dynamic adjustments of compression levels based on real-time factors such as resource availability or network stability.

For instance, if a significant number of devices in a particular area are approaching a critical battery level, the system could increase the compression 
 level to further reduce energy consumption, ensuring that devices can continue participating in the process for longer. 
Conversely, if most devices have sufficient resources but the number of federations fails to stabilize over time, the system could lower the 
 compression level, aiming for a better approximation of the data and improving federation stability. 

\section{Opportunities and Challenges}\label{sec:impact}
\begin{figure}
    \centering
    \begin{subfigure}[b]{0.49\columnwidth}
        \includegraphics[width=\linewidth]{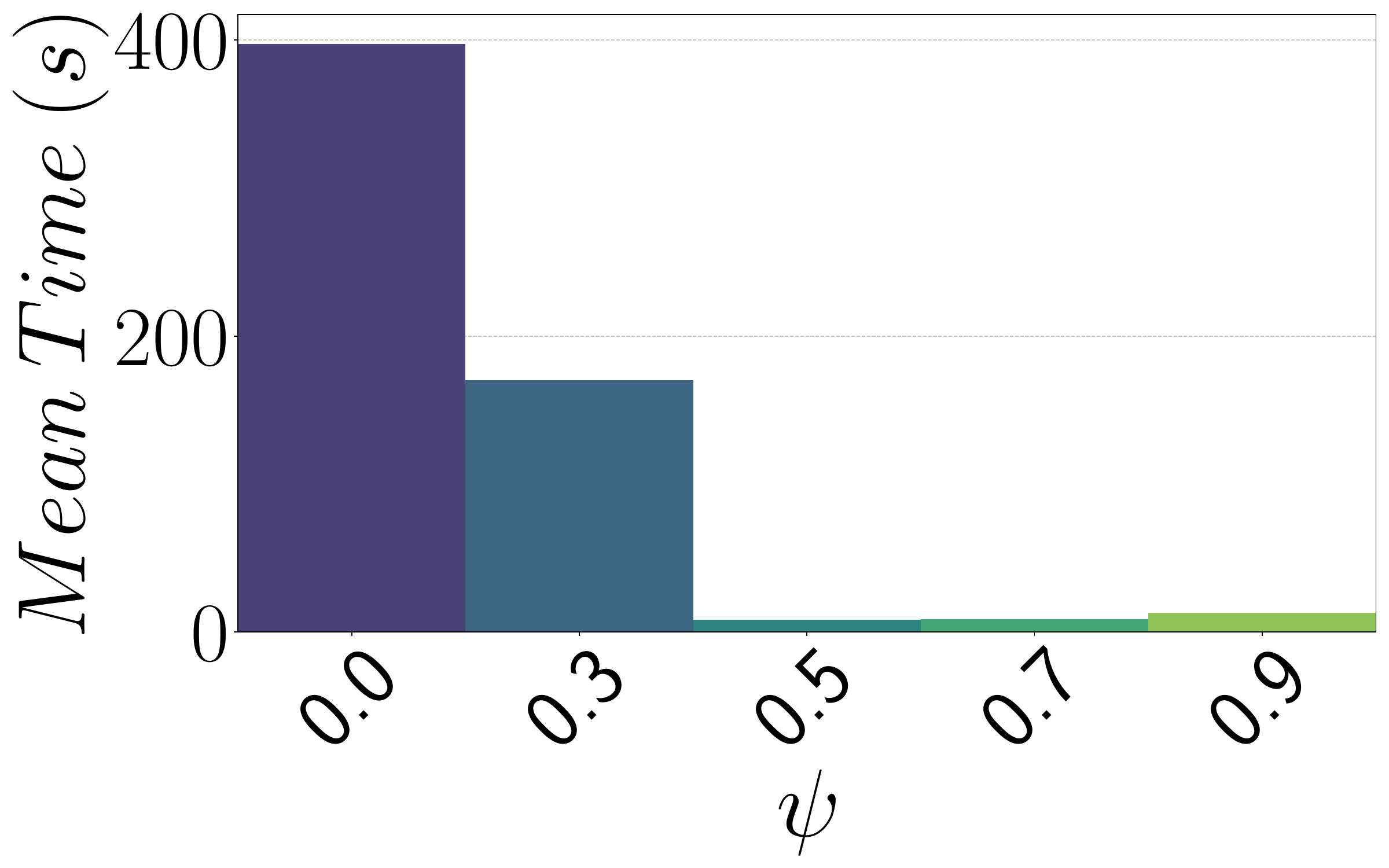}
    \end{subfigure}
    \begin{subfigure}[b]{0.49\columnwidth}
        \includegraphics[width=\linewidth]{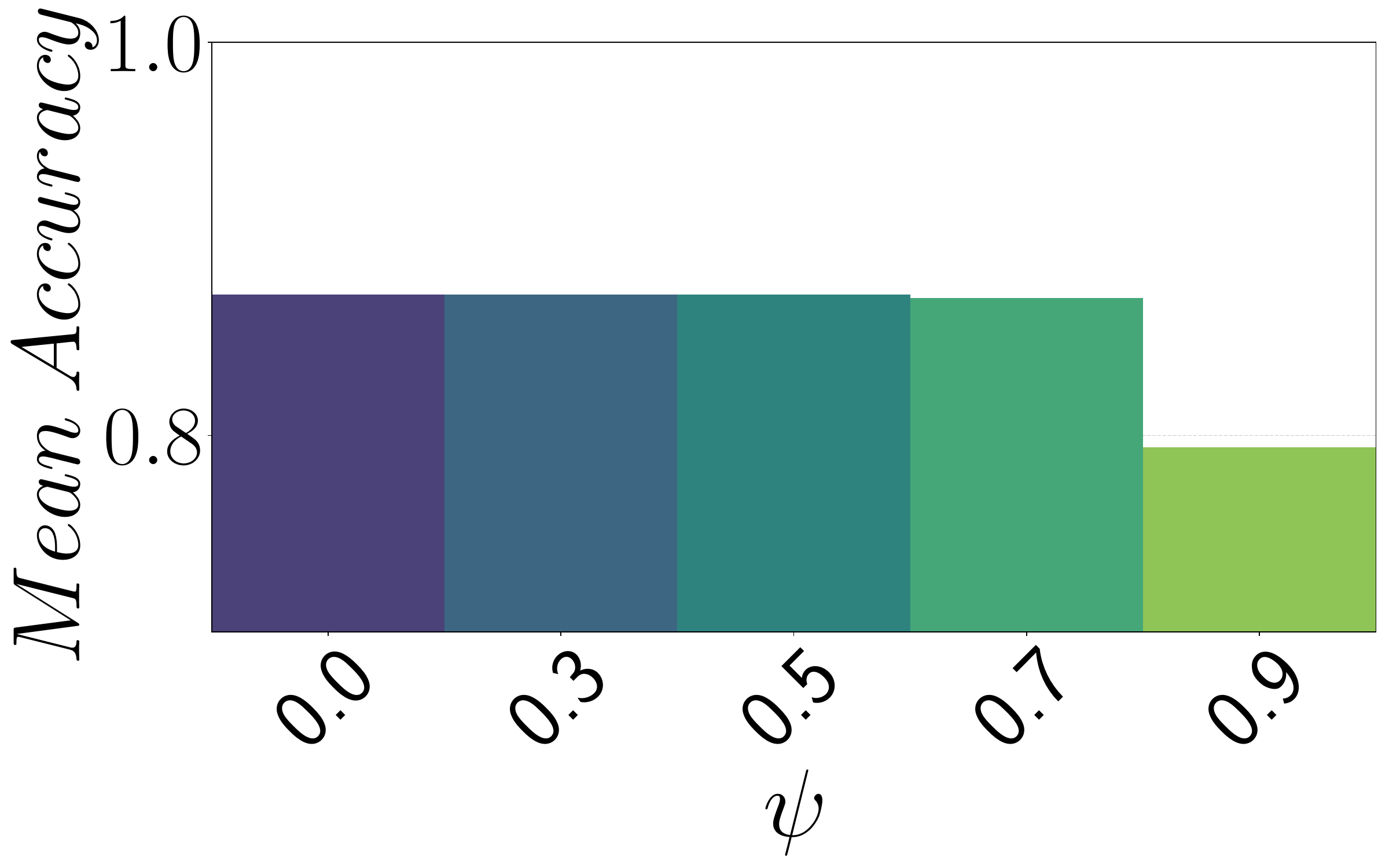}
    \end{subfigure}
    \caption{
    Impact of sparsification level $\psi$ on neural network performance.
    }
    \label{fig:sparsification-comparison}
\end{figure}

Integrating sparsification and quantization will lead to a reduction in inference time and model size, making neural networks more efficient 
 for deployment on resource-constrained devices. 
\Cref{fig:sparsification-comparison} presents the results of an experiment where an MLP network, trained on the Extended MNIST dataset, was subjected to post-pruning at different sparsification levels, namely: $\psi \in \{0.0, 0.3, 0.5, 0.7, 0.9\}$. 
The results highlight a significant reduction in inference time during testing as sparsification increases. 
Notably, pruning 30\% of the weights already reduces inference time by nearly two-thirds, without any noticeable drop in accuracy,
 and cuts energy consumption by about half (\Cref{tab:test}).
A degradation in performance is observed only at extreme sparsification levels (\textit{e.g.}, $\psi = 0.9$).
Similarly, quantization reduces model size, enabling lower memory consumption on devices and a decrease in bandwidth usage when exchanging models
 with neighbors or aggregators. 
In our preliminary experiments, using 8-bit integer quantization, the model size was reduced to approximately one fourth of the original, 
 occupying 454KB compared to 1.8MB.
This is particularly beneficial in decentralized learning settings where communication efficiency is a limiting factor.

Despite these advantages, several challenges must be addressed. 
One major issue arises from the heterogeneity introduced by iterative pruning techniques. 
Since these methods modify the network structure dynamically during training, different devices may end up with 
 architecturally diverse sparse models. 
 This makes aggregation more complex, requiring the development of techniques capable of merging networks with varying 
 sparsity patterns while preserving model integrity.

Another challenge stems from the gap between theoretical expectations and practical implementations of compression frameworks. 
Many compression strategies are still active research topics, and current state-of-the-art implementations may not yield the predicted 
 resource savings in real-world deployments. 
To accurately assess their impact, it is necessary to conduct a comparative study of existing libraries and frameworks, 
 evaluating their actual improvements in terms of battery consumption, inference time, and communication efficiency. 
This, in turn, requires the formalization of evaluation metrics that can provide objective insights into their effectiveness.

Finally, a common issue in federated learning research is the lack of comprehensive benchmarking frameworks. 
While some initial efforts have been made--such as Flower~\cite{DBLP:journals/corr/abs-2007-14390} and 
 ProFed~\cite{DBLP:journals/corr/abs-profed}--these frameworks are still under development and lack the maturity 
 needed for extensive benchmarking. 
Advancing this aspect of the field will be crucial for establishing standardized evaluation methodologies that enable meaningful 
 comparisons between different approaches. 

\begin{table}[t]
    \centering
    \caption{Relative energy consumption of sparse neural networks at different sparsification levels ($\psi$), 
     normalized with respect to the dense baseline ($\psi = 0.0$). 
     Energy consumption has been measured using PyJoules~\cite{Belgaid_Pyjoules_Python_library_2019}.}
    \label{tab:test}
    \begin{tabular}{c|c}
    \toprule
    \textbf{Sparsification ($\psi$)} & \textbf{Relative Consumption} \\ 
    \midrule
    $0.3$        & $41\%$   \\ 
    $0.5$        & $3\%$   \\ 
    $0.7$        & $3\%$   \\ 
    $0.9$        & $4\%$   \\ 
    \bottomrule
    \end{tabular}
\end{table}

\section{Conclusion}\label{sec:conclusion}
In this paper, 
 we presented a vision for more sustainable federated learning approaches aligned with Society 5.0 requirements. 
Specifically, we introduced \approach{}, 
 a federated learning framework designed 
 for large-scale IoT systems that integrates neural network compression techniques 
 to reduce resource consumption of the learning process. 
Through our preliminary experiments, 
 we highlight that \approach{} can reduce power consumption and (partially) 
 communication overhead while maintaining competitive accuracy even with significant sparsification. 

The research roadmap we outlined provides a clear path for implementing and evaluating our approach in real-world settings.
We believe this work lays a foundation for future research 
 in environmentally sustainable federated learning for large-scale IoT ecosystems, 
 contributing to both Society 5.0's human-centered vision and 
 the imperative for resource-efficient distributed machine learning solutions.

\section*{Acknowledgment}
This work is supported by the Italian PRIN project ``CommonWears'' (2020 HCWWLP).

\bibliographystyle{IEEEtran}
\bibliography{IEEEfull}

\begin{thebibliography}{10}
\providecommand{\url}[1]{#1}
\csname url@samestyle\endcsname
\providecommand{\newblock}{\relax}
\providecommand{\bibinfo}[2]{#2}
\providecommand{\BIBentrySTDinterwordspacing}{\spaceskip=0pt\relax}
\providecommand{\BIBentryALTinterwordstretchfactor}{4}
\providecommand{\BIBentryALTinterwordspacing}{\spaceskip=\fontdimen2\font plus
\BIBentryALTinterwordstretchfactor\fontdimen3\font minus \fontdimen4\font\relax}
\providecommand{\BIBforeignlanguage}[2]{{%
\expandafter\ifx\csname l@#1\endcsname\relax
\typeout{** WARNING: IEEEtran.bst: No hyphenation pattern has been}%
\typeout{** loaded for the language `#1'. Using the pattern for}%
\typeout{** the default language instead.}%
\else
\language=\csname l@#1\endcsname
\fi
#2}}
\providecommand{\BIBdecl}{\relax}
\BIBdecl

\bibitem{deguchi2020society}
A.~Deguchi, C.~Hirai, H.~Matsuoka, T.~Nakano, K.~Oshima, M.~Tai, and S.~Tani, ``What is society 5.0,'' \emph{Society}, vol.~5, no.~0, pp. 1--24, 2020.

\bibitem{DBLP:conf/aaai/ConitzerO23}
\BIBentryALTinterwordspacing
V.~Conitzer and C.~Oesterheld, ``Foundations of cooperative {AI},'' in \emph{Thirty-Seventh {AAAI} Conference on Artificial Intelligence, {AAAI} 2023, Thirty-Fifth Conference on Innovative Applications of Artificial Intelligence, {IAAI} 2023, Thirteenth Symposium on Educational Advances in Artificial Intelligence, {EAAI} 2023, Washington, DC, USA, February 7-14, 2023}, B.~Williams, Y.~Chen, and J.~Neville, Eds.\hskip 1em plus 0.5em minus 0.4em\relax {AAAI} Press, 2023, pp. 15\,359--15\,367. [Online]. Available: \url{https://doi.org/10.1609/aaai.v37i13.26791}
\BIBentrySTDinterwordspacing

\bibitem{DBLP:journals/tit/GhoshCYR22}
\BIBentryALTinterwordspacing
A.~Ghosh, J.~Chung, D.~Yin, and K.~Ramchandran, ``An efficient framework for clustered federated learning,'' \emph{{IEEE} Trans. Inf. Theory}, vol.~68, no.~12, pp. 8076--8091, 2022. [Online]. Available: \url{https://doi.org/10.1109/TIT.2022.3192506}
\BIBentrySTDinterwordspacing

\bibitem{DBLP:journals/tbd/LiLLZSLZWW24}
\BIBentryALTinterwordspacing
Z.~Li, J.~Lu, S.~Luo, D.~Zhu, Y.~Shao, Y.~Li, Z.~Zhang, Y.~Wang, and C.~Wu, ``Towards effective clustered federated learning: {A} peer-to-peer framework with adaptive neighbor matching,'' \emph{{IEEE} Trans. Big Data}, vol.~10, no.~6, pp. 812--826, 2024. [Online]. Available: \url{https://doi.org/10.1109/TBDATA.2022.3222971}
\BIBentrySTDinterwordspacing

\bibitem{DBLP:journals/iotj/PrakashDCQSCGP22}
\BIBentryALTinterwordspacing
P.~Prakash, J.~Ding, R.~Chen, X.~Qin, M.~Shu, Q.~Cui, Y.~Guo, and M.~Pan, ``Iot device friendly and communication-efficient federated learning via joint model pruning and quantization,'' \emph{{IEEE} Internet Things J.}, vol.~9, no.~15, pp. 13\,638--13\,650, 2022. [Online]. Available: \url{https://doi.org/10.1109/JIOT.2022.3145865}
\BIBentrySTDinterwordspacing

\bibitem{DBLP:conf/acsos/Domini24}
\BIBentryALTinterwordspacing
D.~Domini, ``Towards self-adaptive cooperative learning in collective systems,'' in \emph{{IEEE} International Conference on Autonomic Computing and Self-Organizing Systems, {ACSOS} 2024 - Companion, Aarhus, Denmark, September 16-20, 2024}.\hskip 1em plus 0.5em minus 0.4em\relax {IEEE}, 2024, pp. 158--160. [Online]. Available: \url{https://doi.org/10.1109/ACSOS-C63493.2024.00049}
\BIBentrySTDinterwordspacing

\bibitem{DBLP:journals/jpdc/HegedusDJ21}
\BIBentryALTinterwordspacing
I.~Heged{\"{u}}s, G.~Danner, and M.~Jelasity, ``Decentralized learning works: An empirical comparison of gossip learning and federated learning,'' \emph{J. Parallel Distributed Comput.}, vol. 148, pp. 109--124, 2021. [Online]. Available: \url{https://doi.org/10.1016/j.jpdc.2020.10.006}
\BIBentrySTDinterwordspacing

\bibitem{DBLP:conf/acsos/DominiAFVE24}
\BIBentryALTinterwordspacing
D.~Domini, G.~Aguzzi, N.~Farabegoli, M.~Viroli, and L.~Esterle, ``Proximity-based self-federated learning,'' in \emph{{IEEE} International Conference on Autonomic Computing and Self-Organizing Systems, {ACSOS} 2024, Aarhus, Denmark, September 16-20, 2024}.\hskip 1em plus 0.5em minus 0.4em\relax {IEEE}, 2024, pp. 139--144. [Online]. Available: \url{https://doi.org/10.1109/ACSOS61780.2024.00033}
\BIBentrySTDinterwordspacing

\bibitem{DBLP:conf/aistats/McMahanMRHA17}
\BIBentryALTinterwordspacing
B.~McMahan, E.~Moore, D.~Ramage, S.~Hampson, and B.~A. y~Arcas, ``Communication-efficient learning of deep networks from decentralized data,'' in \emph{Proceedings of the 20th International Conference on Artificial Intelligence and Statistics, {AISTATS} 2017, 20-22 April 2017, Fort Lauderdale, FL, {USA}}, ser. Proceedings of Machine Learning Research, A.~Singh and X.~J. Zhu, Eds., vol.~54.\hskip 1em plus 0.5em minus 0.4em\relax {PMLR}, 2017, pp. 1273--1282. [Online]. Available: \url{http://proceedings.mlr.press/v54/mcmahan17a.html}
\BIBentrySTDinterwordspacing

\bibitem{DBLP:journals/corr/McMahanMRA16}
\BIBentryALTinterwordspacing
H.~B. McMahan, E.~Moore, D.~Ramage, and B.~A. y~Arcas, ``Federated learning of deep networks using model averaging,'' \emph{CoRR}, vol. abs/1602.05629, 2016. [Online]. Available: \url{http://arxiv.org/abs/1602.05629}
\BIBentrySTDinterwordspacing

\bibitem{DBLP:conf/ijcai/Zhang00XLC24}
\BIBentryALTinterwordspacing
Y.~Zhang, H.~Lu, N.~Liu, Y.~Xu, Q.~Li, and L.~Cui, ``Personalized federated learning for cross-city traffic prediction,'' in \emph{Proceedings of the Thirty-Third International Joint Conference on Artificial Intelligence, {IJCAI} 2024, Jeju, South Korea, August 3-9, 2024}.\hskip 1em plus 0.5em minus 0.4em\relax ijcai.org, 2024, pp. 5526--5534. [Online]. Available: \url{https://www.ijcai.org/proceedings/2024/611}
\BIBentrySTDinterwordspacing

\bibitem{DBLP:journals/corr/abs-2403-07444}
\BIBentryALTinterwordspacing
R.~Zhang, H.~Wang, B.~Li, X.~Cheng, and L.~Yang, ``A survey on federated learning in intelligent transportation systems,'' \emph{CoRR}, vol. abs/2403.07444, 2024. [Online]. Available: \url{https://doi.org/10.48550/arXiv.2403.07444}
\BIBentrySTDinterwordspacing

\bibitem{DBLP:journals/iotj/ImteajTWLA22}
\BIBentryALTinterwordspacing
A.~Imteaj, U.~Thakker, S.~Wang, J.~Li, and M.~H. Amini, ``A survey on federated learning for resource-constrained iot devices,'' \emph{{IEEE} Internet Things J.}, vol.~9, no.~1, pp. 1--24, 2022. [Online]. Available: \url{https://doi.org/10.1109/JIOT.2021.3095077}
\BIBentrySTDinterwordspacing

\bibitem{DBLP:conf/icde/LiDCH22}
\BIBentryALTinterwordspacing
Q.~Li, Y.~Diao, Q.~Chen, and B.~He, ``Federated learning on non-iid data silos: An experimental study,'' in \emph{Proceedings of the International Conference on Data Engineering}.\hskip 1em plus 0.5em minus 0.4em\relax {IEEE}, 2022, pp. 965--978. [Online]. Available: \url{https://doi.org/10.1109/ICDE53745.2022.00077}
\BIBentrySTDinterwordspacing

\bibitem{Imteaj2023}
\BIBentryALTinterwordspacing
A.~Imteaj, K.~Mamun~Ahmed, U.~Thakker, S.~Wang, J.~Li, and M.~H. Amini, \emph{Federated Learning for Resource-Constrained IoT Devices: Panoramas and State of the Art}.\hskip 1em plus 0.5em minus 0.4em\relax Cham: Springer International Publishing, 2023, pp. 7--27. [Online]. Available: \url{https://doi.org/10.1007/978-3-031-11748-0\_2}
\BIBentrySTDinterwordspacing

\bibitem{DBLP:conf/mlsys/LiSZSTS20}
T.~Li, A.~K. Sahu, M.~Zaheer, M.~Sanjabi, A.~Talwalkar, and V.~Smith, ``Federated optimization in heterogeneous networks,'' in \emph{Proceedings of the Third Conference on Machine Learning and Systems, MLSys 2020, Austin, TX, USA, March 2-4, 2020}, I.~S. Dhillon, D.~S. Papailiopoulos, and V.~Sze, Eds.\hskip 1em plus 0.5em minus 0.4em\relax mlsys.org, 2020.

\bibitem{DBLP:conf/icml/KarimireddyKMRS20}
\BIBentryALTinterwordspacing
S.~P. Karimireddy, S.~Kale, M.~Mohri, S.~J. Reddi, S.~U. Stich, and A.~T. Suresh, ``{SCAFFOLD:} stochastic controlled averaging for federated learning,'' in \emph{Proceedings of the 37th International Conference on Machine Learning, {ICML} 2020, 13-18 July 2020, Virtual Event}, ser. Proceedings of Machine Learning Research, vol. 119.\hskip 1em plus 0.5em minus 0.4em\relax {PMLR}, 2020, pp. 5132--5143. [Online]. Available: \url{http://proceedings.mlr.press/v119/karimireddy20a.html}
\BIBentrySTDinterwordspacing

\bibitem{DBLP:conf/ecai/Li0TWXZ23}
\BIBentryALTinterwordspacing
X.~Li, X.~Chen, B.~Tang, S.~Wang, Y.~Xuan, and Z.~Zhao, ``Unsupervised graph structure-assisted personalized federated learning,'' in \emph{Proceedings of the European Conference on Artificial Intelligence}, ser. Frontiers in Artificial Intelligence and Applications, vol. 372.\hskip 1em plus 0.5em minus 0.4em\relax {IOS} Press, 2023, pp. 1430--1438. [Online]. Available: \url{https://doi.org/10.3233/FAIA230421}
\BIBentrySTDinterwordspacing

\bibitem{esterle2022deep}
L.~Esterle, ``Deep learning in multiagent systems,'' in \emph{Deep Learning for Robot Perception and Cognition}.\hskip 1em plus 0.5em minus 0.4em\relax Elsevier, 2022, pp. 435--460.

\bibitem{DBLP:conf/sac/MalucelliDAV25}
N.~Malucelli, D.~Domini, G.~Aguzzi, and M.~Viroli, ``Neighbor-based decentralized training strategies for multi-agent reinforcement learning,'' in \emph{Proceedings of the 40th {ACM/SIGAPP} Symposium on Applied Computing, {SAC} 2025, Catania, Italy, March 31-April 4, 2025}.\hskip 1em plus 0.5em minus 0.4em\relax {ACM}, 2024, pp. 3--10.

\bibitem{DBLP:conf/coordination/DominiAEV24}
\BIBentryALTinterwordspacing
D.~Domini, G.~Aguzzi, L.~Esterle, and M.~Viroli, ``Field-based coordination for federated learning,'' in \emph{Coordination Models and Languages - 26th {IFIP} {WG} 6.1 International Conference, {COORDINATION} 2024, Held as Part of the 19th International Federated Conference on Distributed Computing Techniques, DisCoTec 2024, Groningen, The Netherlands, June 17-21, 2024, Proceedings}, ser. Lecture Notes in Computer Science, I.~Castellani and F.~Tiezzi, Eds., vol. 14676.\hskip 1em plus 0.5em minus 0.4em\relax Springer, 2024, pp. 56--74. [Online]. Available: \url{https://doi.org/10.1007/978-3-031-62697-5\_4}
\BIBentrySTDinterwordspacing

\bibitem{Li2020-xw}
E.~Li, L.~Zeng, Z.~Zhou, and X.~Chen, ``Edge {AI}: On-demand accelerating deep neural network inference via edge computing,'' \emph{IEEE Trans. Wirel. Commun.}, vol.~19, no.~1, pp. 447--457, Jan. 2020.

\bibitem{Liang2021-ta}
T.~Liang, J.~Glossner, L.~Wang, S.~Shi, and X.~Zhang, ``Pruning and quantization for deep neural network acceleration: A survey,'' \emph{Neurocomputing}, vol. 461, pp. 370--403, Oct. 2021.

\bibitem{Erhan2025-gi}
L.~Erhan, A.~Liotta, and L.~Cavallaro, ``Comparing training of sparse to classic neural networks for binary classification in medical data,'' in \emph{Lecture Notes in Computer Science}, ser. Lecture notes in computer science.\hskip 1em plus 0.5em minus 0.4em\relax Cham: Springer Nature Switzerland, 2025, pp. 101--106.

\bibitem{Jayasimhan2024-mq}
\BIBentryALTinterwordspacing
A.~Jayasimhan and P.~P., ``Resprune: An energy-efficient restorative filter pruning method using stochastic optimization for accelerating cnn,'' \emph{Pattern Recognition}, vol. 155, p. 110671, Nov 2024. [Online]. Available: \url{https://www.sciencedirect.com/science/article/pii/S0031320324004229}
\BIBentrySTDinterwordspacing

\bibitem{Atashgahi2024-ki}
\BIBentryALTinterwordspacing
Z.~Atashgahi, T.~Liu, M.~Pechenizkiy, R.~Veldhuis, D.~C. Mocanu, and M.~van~der Schaar, ``Unveiling the power of sparse neural networks for feature selection,'' Oct. 2024. [Online]. Available: \url{https://arxiv.org/abs/2408.04583}
\BIBentrySTDinterwordspacing

\bibitem{krishnamoorthi2018quantizingdeepconvolutionalnetworks}
\BIBentryALTinterwordspacing
R.~Krishnamoorthi, ``Quantizing deep convolutional networks for efficient inference: A whitepaper,'' 2018. [Online]. Available: \url{https://arxiv.org/abs/1806.08342}
\BIBentrySTDinterwordspacing

\bibitem{DBLP:journals/computer/BealPV15}
\BIBentryALTinterwordspacing
J.~Beal, D.~Pianini, and M.~Viroli, ``Aggregate programming for the internet of things,'' \emph{Computer}, vol.~48, no.~9, pp. 22--30, 2015. [Online]. Available: \url{https://doi.org/10.1109/MC.2015.261}
\BIBentrySTDinterwordspacing

\bibitem{DBLP:journals/csur/Casadei23}
\BIBentryALTinterwordspacing
R.~Casadei, ``Macroprogramming: Concepts, state of the art, and opportunities of macroscopic behaviour modelling,'' \emph{{ACM} Comput. Surv.}, vol.~55, no. 13s, pp. 275:1--275:37, 2023. [Online]. Available: \url{https://doi.org/10.1145/3579353}
\BIBentrySTDinterwordspacing

\bibitem{DBLP:journals/pervasive/MameiZL04}
\BIBentryALTinterwordspacing
M.~Mamei, F.~Zambonelli, and L.~Leonardi, ``Co-fields: {A} physically inspired approach to motion coordination,'' \emph{{IEEE} Pervasive Comput.}, vol.~3, no.~2, pp. 52--61, 2004. [Online]. Available: \url{https://doi.org/10.1109/MPRV.2004.1316820}
\BIBentrySTDinterwordspacing

\bibitem{DBLP:journals/tocl/AudritoVDPB19}
\BIBentryALTinterwordspacing
G.~Audrito, M.~Viroli, F.~Damiani, D.~Pianini, and J.~Beal, ``A higher-order calculus of computational fields,'' \emph{{ACM} Trans. Comput. Log.}, vol.~20, no.~1, pp. 5:1--5:55, 2019. [Online]. Available: \url{https://doi.org/10.1145/3285956}
\BIBentrySTDinterwordspacing

\bibitem{DBLP:conf/coordination/AudritoBDV18}
\BIBentryALTinterwordspacing
G.~Audrito, J.~Beal, F.~Damiani, and M.~Viroli, ``Space-time universality of field calculus,'' in \emph{Coordination Models and Languages - 20th {IFIP} {WG} 6.1 International Conference, {COORDINATION} 2018, Held as Part of the 13th International Federated Conference on Distributed Computing Techniques, DisCoTec 2018, Madrid, Spain, June 18-21, 2018. Proceedings}, ser. Lecture Notes in Computer Science, G.~D.~M. Serugendo and M.~Loreti, Eds., vol. 10852.\hskip 1em plus 0.5em minus 0.4em\relax Springer, 2018, pp. 1--20. [Online]. Available: \url{https://doi.org/10.1007/978-3-319-92408-3\_1}
\BIBentrySTDinterwordspacing

\bibitem{DBLP:journals/tomacs/ViroliABDP18}
\BIBentryALTinterwordspacing
M.~Viroli, G.~Audrito, J.~Beal, F.~Damiani, and D.~Pianini, ``Engineering resilient collective adaptive systems by self-stabilisation,'' \emph{{ACM} Trans. Model. Comput. Simul.}, vol.~28, no.~2, pp. 16:1--16:28, 2018. [Online]. Available: \url{https://doi.org/10.1145/3177774}
\BIBentrySTDinterwordspacing

\bibitem{DBLP:journals/taas/BealVPD17}
\BIBentryALTinterwordspacing
J.~Beal, M.~Viroli, D.~Pianini, and F.~Damiani, ``Self-adaptation to device distribution in the internet of things,'' \emph{{ACM} Trans. Auton. Adapt. Syst.}, vol.~12, no.~3, pp. 12:1--12:29, 2017. [Online]. Available: \url{https://doi.org/10.1145/3105758}
\BIBentrySTDinterwordspacing

\bibitem{DBLP:conf/coordination/CasadeiPVN19}
\BIBentryALTinterwordspacing
R.~Casadei, D.~Pianini, M.~Viroli, and A.~Natali, ``Self-organising coordination regions: {A} pattern for edge computing,'' in \emph{Coordination Models and Languages - 21st {IFIP} {WG} 6.1 International Conference, {COORDINATION} 2019, Held as Part of the 14th International Federated Conference on Distributed Computing Techniques, DisCoTec 2019, Kongens Lyngby, Denmark, June 17-21, 2019, Proceedings}, ser. Lecture Notes in Computer Science, H.~R. Nielson and E.~Tuosto, Eds., vol. 11533.\hskip 1em plus 0.5em minus 0.4em\relax Springer, 2019, pp. 182--199. [Online]. Available: \url{https://doi.org/10.1007/978-3-030-22397-7\_11}
\BIBentrySTDinterwordspacing

\bibitem{DBLP:journals/corr/abs-2502-08577}
\BIBentryALTinterwordspacing
D.~Domini, G.~Aguzzi, L.~Esterle, and M.~Viroli, ``{FBFL:} {A} field-based coordination approach for data heterogeneity in federated learning,'' \emph{CoRR}, vol. abs/2502.08577, 2025. [Online]. Available: \url{https://doi.org/10.48550/arXiv.2502.08577}
\BIBentrySTDinterwordspacing

\bibitem{DBLP:journals/scp/DominiCAV24}
\BIBentryALTinterwordspacing
D.~Domini, F.~Cavallari, G.~Aguzzi, and M.~Viroli, ``Scarlib: Towards a hybrid toolchain for aggregate computing and many-agent reinforcement learning,'' \emph{Sci. Comput. Program.}, vol. 238, p. 103176, 2024. [Online]. Available: \url{https://doi.org/10.1016/j.scico.2024.103176}
\BIBentrySTDinterwordspacing

\bibitem{DBLP:conf/acsos/AguzziVE23}
\BIBentryALTinterwordspacing
G.~Aguzzi, M.~Viroli, and L.~Esterle, ``Field-informed reinforcement learning of collective tasks with graph neural networks,'' in \emph{{IEEE} International Conference on Autonomic Computing and Self-Organizing Systems, {ACSOS} 2023, Toronto, ON, Canada, September 25-29, 2023}.\hskip 1em plus 0.5em minus 0.4em\relax {IEEE}, 2023, pp. 37--46. [Online]. Available: \url{https://doi.org/10.1109/ACSOS58161.2023.00021}
\BIBentrySTDinterwordspacing

\bibitem{mocanu2018scalable}
D.~C. Mocanu, E.~Mocanu, P.~Stone, P.~H. Nguyen, M.~Gibescu, and A.~Liotta, ``Scalable training of artificial neural networks with adaptive sparse connectivity inspired by network science,'' \emph{Nature communications}, vol.~9, no.~1, p. 2383, 2018.

\bibitem{lecun2010mnist}
Y.~LeCun, C.~Cortes, C.~Burges \emph{et~al.}, ``Mnist handwritten digit database,'' 2010.

\bibitem{cifar10}
\BIBentryALTinterwordspacing
A.~Krizhevsky, V.~Nair, and G.~Hinton, ``Cifar-10 (canadian institute for advanced research).'' [Online]. Available: \url{http://www.cs.toronto.edu/~kriz/cifar.html}
\BIBentrySTDinterwordspacing

\bibitem{DBLP:journals/corr/abs-2007-14390}
\BIBentryALTinterwordspacing
D.~J. Beutel, T.~Topal, A.~Mathur, X.~Qiu, T.~Parcollet, and N.~D. Lane, ``Flower: {A} friendly federated learning research framework,'' \emph{CoRR}, vol. abs/2007.14390, 2020. [Online]. Available: \url{https://arxiv.org/abs/2007.14390}
\BIBentrySTDinterwordspacing

\bibitem{DBLP:journals/corr/abs-profed}
\BIBentryALTinterwordspacing
D.~Domini, G.~Aguzzi, and M.~Viroli, ``Profed: a benchmark for proximity-based non-iid federated learning,'' \emph{CoRR}, vol. abs/2503.20618, 2025. [Online]. Available: \url{https://arxiv.org/abs/2503.20618}
\BIBentrySTDinterwordspacing

\bibitem{Belgaid_Pyjoules_Python_library_2019}
\BIBentryALTinterwordspacing
M.~c. Belgaid, R.~Rouvoy, and L.~Seinturier, ``{Pyjoules: Python library that measures python code snippets},'' Nov. 2019. [Online]. Available: \url{https://github.com/powerapi-ng/pyJoules}
\BIBentrySTDinterwordspacing

\end{thebibliography}

\end{document}